\documentclass[10pt,twocolumn,letterpaper]{article}

\usepackage{cvpr}              
\usepackage{multirow}
\usepackage{booktabs, xcolor, colortbl, arydshln, tikz}
\usepackage[most]{tcolorbox}
\usepackage{algorithm}
\usepackage{algorithmic}

\newcommand\mypara[1]{\noindent\textbf{#1.}}

\newcommand{\ourmethod}{HorizonForge\xspace}
\newcommand{\ourbenchmark}{HorizonSuite\xspace}

\newcommand{\ourtitle}{HorizonForge: Driving Scene Editing with Any Trajectories and Any Vehicles}

\definecolor{cvprblue}{rgb}{0.21,0.49,0.74}
\usepackage[pagebackref,breaklinks,colorlinks,allcolors=cvprblue]{hyperref}

\def\confName{CVPR}
\def\confYear{2025}

\title{\ourtitle }

\author{%
  Yifan Wang$^{1\, 2}$ \quad 
  Francesco Pittaluga$^1$ \quad 
  Zaid Tasneem$^1$ \\ 
  Chenyu You$^2$ \quad 
  Manmohan Chandraker$^{1\, 3}$ \quad 
  Ziyu Jiang$^1$ \\ \\
$^{1}$NEC Labs America \quad
$^{2}$Stony Brook University \quad
$^{3}$UC San Diego 
}

\begin{document}

\twocolumn[{
\renewcommand\twocolumn[1][]{#1}
\maketitle

\begin{center}
    \centering
    \captionsetup{type=figure}
    \includegraphics[width=1.0\linewidth]
    {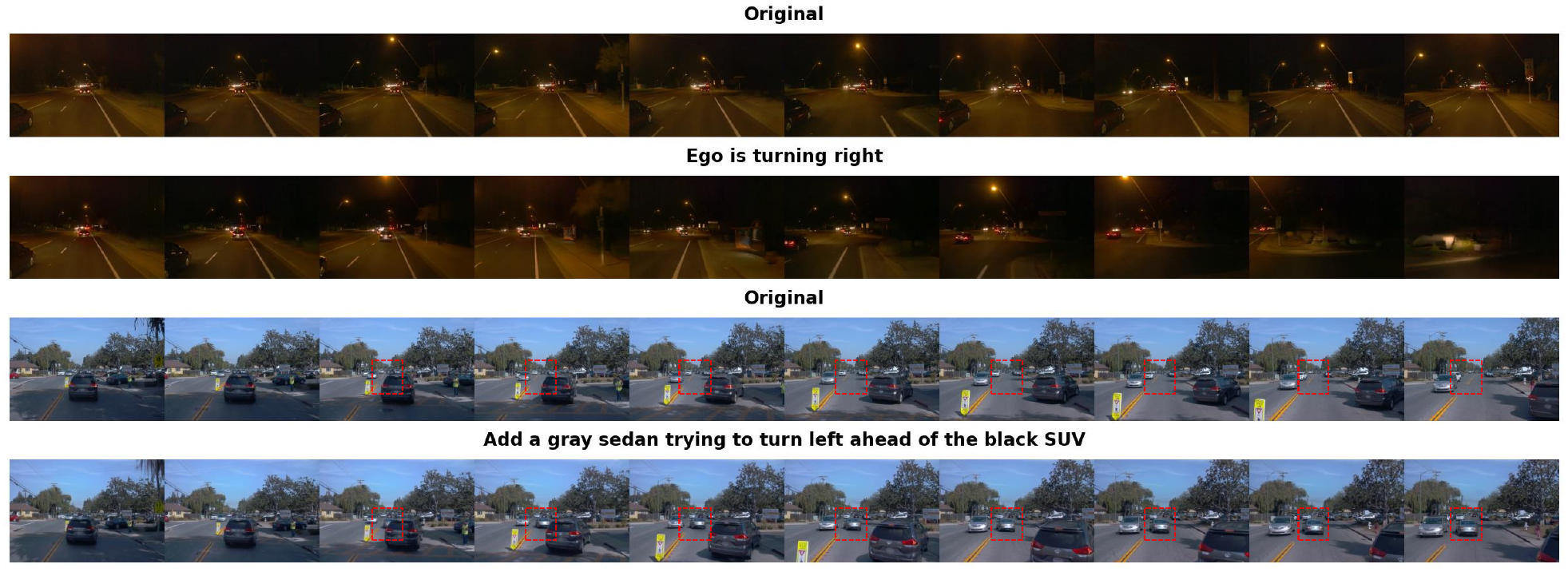}
    \captionof{figure}{
    \textbf{\ourmethod} is capable of generating high-quality driving scenes in accordance with the provided manipulation instructions. The top two rows of the image depict the transformation of the ego car to the right, while the bottom two rows illustrate the insertion of a gray sedan in front of the selected SUV at the red box location.
    }
    \label{fig:teaser}
\end{center}
}]

\begin{abstract}
\vspace{-20pt}

Controllable driving scene generation is critical for realistic and scalable autonomous driving simulation, yet existing approaches struggle to jointly achieve photorealism and precise control. We introduce \textbf{\ourmethod}, a unified framework that reconstructs scenes as editable Gaussian Splats and Meshes, enabling fine-grained 3D manipulation and language-driven vehicle insertion. Edits are rendered through a noise-aware video diffusion process that enforces spatial and temporal consistency, producing diverse scene variations in a single feed-forward pass without per-trajectory optimization. To standardize evaluation, we further propose \textbf{\ourbenchmark}, a comprehensive benchmark spanning ego- and agent-level editing tasks such as trajectory modifications and object manipulation. Extensive experiments show that Gaussian-Mesh representation delivers substantially higher fidelity than alternative 3D representations, and that temporal priors from video diffusion are essential for coherent synthesis. Combining these findings, \textbf{\ourmethod} establishes a simple yet powerful paradigm for photorealistic, controllable driving simulation, achieving an 83.4\% user-preference gain and a 25.19\% FID improvement over the second best state-of-the-art method. Project page: \url{https://horizonforge.github.io/}.

\end{abstract}    
\vspace{-10pt}
\section{Introduction}
\label{sec:intro}

Evaluating autonomous driving algorithms under long-tail scenarios is essential for ensuring safety during real-world deployment. Although modern autonomous driving systems collect vast amounts of daily driving logs, rare but safety-critical events, such as aggressive lane-switching or sudden braking, remain extremely difficult and expensive to capture at scale. Recent advances in simulation have made it increasingly feasible to synthesize such scenarios by editing existing driving logs. However, achieving both high photorealism and precise controllability remains challenging. Reconstruction-based methods~\cite{kerbl3dgaussians, chen2024omnire} offer explicit geometric fidelity through accurate 3D modeling, yet often fail to generalize to unseen regions~\cite{zhang2022ray}. In contrast, generation-based approaches can hallucinate new content but frequently struggle to preserve scene structure and appearance, limiting their ability to perform fine-grained traffic editing~\cite{gao2024magicdrive3d}. Although recent hybrid methods~\cite{ni2025recondreamer,zhao2024drivedreamer4d} have shown promising results, a simple, unified framework coupled with a systematic study of controllable driving editing is still missing.

To bridge this gap, we propose \textbf{\ourmethod}, a simple yet unified framework for photorealistic and controllable driving scene generation. Our approach begins by harvesting the input scene into editable 3D representations, such as Gaussian Splats and Meshes, that support flexible manipulation in 3D space. Each edit (e.g., lane changes, cut-ins/outs, or vehicle insertion) is then rendered back into the image domain through a noise-aware video diffusion process, ensuring strong spatial and temporal consistency. Additionally, \textbf{\ourmethod} supports language-guided object insertion, enabling the placement of new vehicles described via natural language by generating the corresponding 3D meshes. Once the 3D structure is acquired, diverse controllable variations can be produced in a single feed-forward pass, avoiding the costly per-trajectory optimization required by prior methods~\cite{ni2025recondreamer,zhao2024drivedreamer4d}. Despite its simplicity, \textbf{\ourmethod} achieves state-of-the-art performance across a wide range of editing tasks.

To systematically evaluate controllable editing at scale, we introduce \textbf{\ourbenchmark}, a comprehensive benchmark for controllable driving simulation. It spans a rich set of challenging tasks involving both ego-vehicle and surrounding-agent manipulation, including lane changes, turns, cut-ins/outs, sudden braking or acceleration, and object insertion or removal. This benchmark establishes a unified protocol for assessing fine-grained controllability and visual fidelity across diverse editing scenarios.
Through extensive analysis, we identify two crucial design principles for controllable scene generation:
\begin{enumerate}
    \item \textbf{3D representations matter.} Comparing bounding boxes, colored LiDAR point clouds, and Gaussian Splats (all combined with object meshes), we find that Gaussian Splats encode far richer appearance cues, enabling more accurate edits and substantially improving generation quality.
    \item \textbf{Temporal priors matter.} Our comparison between image- and video-based paradigms shows that video diffusion models more effectively leverage temporal continuity, leading to significantly more coherent and stable scene synthesis.
\end{enumerate}
In summary, our main contributions are 
\begin{itemize}
    \item \textbf{A simple and unified framework (\ourmethod)} for photorealistic and controllable driving scene generation, supporting both 3D editing and language-guided object insertion.
    \item \textbf{A mesh harvesting and insertion pipeline} that enables insertion of any vehicles with only text prompts.
    \item \textbf{A comprehensive benchmark (\ourbenchmark)} enabling standardized evaluation across diverse ego- and agent-level manipulation tasks with fine-grained metrics.
    \item \textbf{Extensive experiments and analysis} showing that \ourmethod\ outperforms state-of-the-art approaches, delivering an 83.4\% user-preference gain and a 25.19\% FID improvement over the second-best method, while offering insights into the key design choices for controllable scene generation.
\end{itemize}

\begin{figure*}[t]
  \centering
  \includegraphics[width=0.9\textwidth]
  {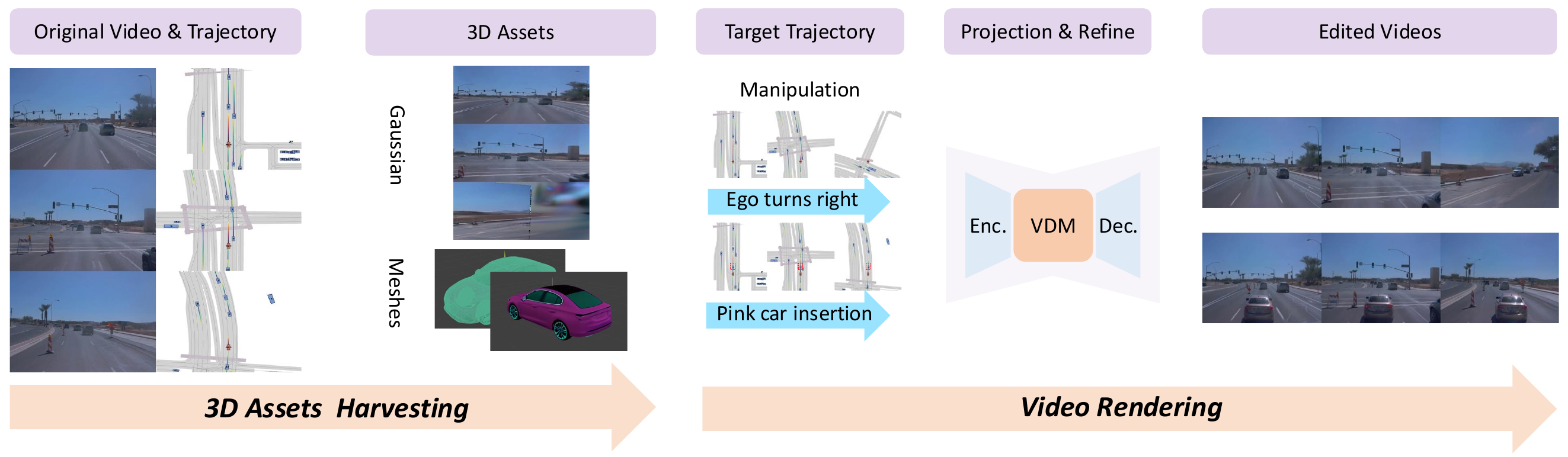}
  \vspace{-10pt}
  \caption{Overview of the \textbf{\ourmethod} framework. With original video and trajectory, we will firstly extract corresponding 3D assets according to the manipulated novel trajectories, then feed them into our rendering model for final generation results.
  }
  \label{fig:pipeline}
  \vspace{-15pt}
\end{figure*}

\section{Related Works}
\subsection{Reconstruction-Based Simulation}
Reconstruction-based simulation methods focus on faithfully recovering the geometry and appearance of real driving scenes for realistic replay or manipulation. 
Large-scale benchmarks such as Waymo~\cite{sun2020waymo}, nuScenes~\cite{caesar2020nuscenes}, and nuPlan~\cite{caesar2021nuplan} provide the empirical foundation for these efforts. 
Recent advances in neural rendering~\cite{
ost2021nsg,tancik2022blocknerf,wu2023mars,xie2023snerf,yang2023emernerf,tonderski2024neurad,yang2023unisim,sun2024lidarf}, particularly 3D Gaussian Splatting~\cite{kerbl3dgaussians, chen2024omnire, yan2024street}, have enabled photorealistic multi-view reconstructions in driving scenarios. 
While these approaches achieve strong visual fidelity and geometric accuracy, their controllability is limited because they primarily replay reconstructed scenes instead of synthesizing novel trajectories or new agents. 

\subsection{Generation-Based Simulation}

In contrast, generation-based methods aim to synthesize controllable driving scenes directly from high-level inputs such as text or semantic maps. 
Diffusion-based generative models~\cite{ho2020ddpm, song2021ddim, ho2022vdm, rombach2022ldm, peebles2023scalable, ho2023imagenvideo, sun2025ouroboros, blattmann2023svd, cheng2023sdfusion, chen2024videocrafter2} have significantly advanced this direction by improving fidelity and temporal coherence, yet most rely on complex multimodal conditioning and 2D cues without explicit physical grounding. 
As a result, their outputs often exhibit stochastic motion and structural inconsistency, making them unreliable for safety-critical domains such as autonomous driving. 
Although recent works~\cite{lu2024fit, zhang2023i2vgen, blattmann2023alignlatents_video, alhaija2025cosmos} improve temporal transfer and latent alignment, they still depend heavily on intricate conditioning pipelines and prompt engineering. 
In this work, we revisit the simplicity of classical video diffusion models, showing that with minimal yet physically grounded conditioning, specifically, trajectory-based control combined with 3D Gaussian and mesh representations, one can achieve both strong generalization and physically consistent controllable generation.

\subsection{Hybrid Simulation}

Hybrid simulation methods seek to combine the geometric accuracy of reconstruction with the flexibility of generative modeling by integrating 3D structural priors into diffusion or transformer frameworks~\cite{gao2023magicdrive, gao2024vista, liu2024mvpbev, yan2025streetcrafter, sun2024coma, hassan2025gem, wu2025difix3d, li2025uniscene, bogdoll2024muvo, you2025uncovering, you2024calibrating, ni2025recondreamer, wang2024drivedreamer, zhao2025drivedreamer,chen2025autoscape, wei2024editable, 
li2025realistic, 
park2025simsplat, 
lu2025infinicube, 
wang2024freevs, 
liang2025driveeditor}. 
While these approaches improve realism and controllability, they often rely on complex architectures and per-trajectory optimization~\cite{ni2025recondreamer,zhao2024drivedreamer4d}, leading to high computational cost and limited scalability. 
In contrast, \textbf{\ourmethod} adopts a simple yet effective paradigm: it reconstructs the scene into editable 3D Gaussian splats and meshes, and employs a standard video diffusion model to render temporally coherent results directly from trajectory-based control. 
This streamlined design eliminates redundant conditioning and heavy optimization while achieving superior photorealism and multi-agent controllability. 
Together with \textbf{\ourbenchmark}, it offers a unified and scalable foundation for physically grounded driving simulation.

\section{\ourmethod}

\subsection{Overview}
\label{subsec:overview}

As illustrated in Figure~\ref{fig:pipeline}, \textbf{\ourmethod} is a unified framework for controllable driving scene generation.
Given a user-specified trajectory set $\mathcal{T} = \{\tau_i\}_{i=1}^N$ that jointly specifies the motion of all vehicles in the scene—including both the ego vehicle and surrounding agents—our system synthesizes a photorealistic and physically consistent driving video through two major stages:

\begin{enumerate}
\item \textbf{3D Assets Harvesting.}
We first reconstruct the input video into a collection of editable 3D assets that can be flexibly manipulated under arbitrary motions in 3D space. To preserve both geometric structure and appearance fidelity, we extract high-quality 3D Gaussian Splat representations~\cite{chen2024omnire}. Through a systematic comparison with alternative modalities, including colored point clouds~\cite{yan2025streetcrafter} and 3D bounding boxes, we find that Gaussian Splats consistently offer superior controllability and rendering quality, owing to their substantially richer appearance detail (Section~\ref{ablation}). 
In addition, we harvest 3D meshes from the input video to support text-driven novel object insertion.  
Specifically, the rendering model is trained with these meshes to enable realistic, geometry-aligned reconstruction of newly inserted vehicles.  
Combined with a text-conditioned mesh generation pipeline (Section~\ref{subsec:3dassets}), this allows \textbf{\ourmethod} to introduce new vehicles described purely in natural language while maintaining scene structure and compatibility with downstream rendering.

\item \textbf{Video Rendering:}
After repositioning all harvested 3D assets according to the target trajectories $\mathcal{T}$, the edited 3D scene is rendered into 2D view sequences. These rasterized frames are subsequently refined by a Video Diffusion Model (VDM), which enforces spatial–temporal consistency and synthesizes high-fidelity driving videos aligned with the specified multi-agent motions. This stage yields photorealistic and temporally stable outputs that faithfully preserve the scene geometry and adhere to the dynamics dictated by $\mathcal{T}$. In Section~\ref{ablation}, we systematically studied the image and video diffusion model and reveal that video diffusion model provides significant boost on the temporal consistency and helps the model to better resolving the artifacts in the rasterized frames.

\end{enumerate}
This paradigm enables \textbf{\ourmethod} to seamlessly integrate explicit multi-agent control with implicit generative modeling, allowing flexible scene manipulation in which the trajectories of all agents, including the ego vehicle, collectively define the resulting dynamics. The design maintains both structural consistency and photorealistic fidelity. Moreover, once the 3D scene representations are harvested, each edited variation can be rendered efficiently with a single feed-forward pass.

In the following sections, we first introduce the necessary preliminaries in Section~\ref{subsec:prelim}. We then describe the 3D Asset Harvesting and Video Rendering components in Section~\ref{subsec:3dassets} and Section~\ref{subsec:video_rendering}, respectively. Finally, we present the proposed benchmark \textbf{\ourbenchmark} in Section~\ref{subsec:benchmark}.

\subsection{Preliminary}
\label{subsec:prelim}

\subsubsection{3D Representation Generation}
\label{sec:prelim_3d}
Our framework leverages explicit 3D representations as conditions to achieve high-fidelity and robust control. We provide a brief overview of 3D Gaussian Splatting.

\textbf{3D Gaussian Splatting (3DGS).}
3DGS~\cite{kerbl3dgaussians} is a high-fidelity, explicit 3D scene representation that enables real-time, photorealistic rendering. A scene is represented by a set of $N$ anisotropic 3D Gaussians. Each Gaussian $G_i$ is defined by following four components,
Position (Mean): $\mu_i \in \mathbb{R}^3$; 
Covariance: $\mathbf{\Sigma}_i \in \mathbb{R}^{3 \times 3}$, typically parameterized by a scale vector $s_i \in \mathbb{R}^3$ and a rotation quaternion $q_i \in \mathbb{R}^4$; 
Opacity: $o_i \in \mathbb{R}$ and 
Color: A set of Spherical Harmonic (SH) coefficients $c_i$ to model view-dependent color.

To render an image from a camera pose, these 3D Gaussians are projected onto the 2D image plane. The final pixel color $C$ is computed by alpha-blending the sorted 2D Gaussians that overlap the pixel in a front-to-back order:
$
C = \sum_{i=1}^N c_i \alpha_i \prod_{j=1}^{i-1} (1 - \alpha_j)
$, 
where $\alpha_i$ is the computed opacity of the $i$-th Gaussian in 2D. 3DGS is known for its ability to capture fine appearance details but can produce noisy artifacts under large, novel view shifts.

\subsubsection{Video Diffusion Models}
\label{sec:prelim_vdm}

Video Diffusion Models~\cite{ho2022vdm} consist of two main components: a forward process and a learned reverse process.

\textbf{Forward Process.}
The forward  process $q$ gradually adds noise to a clean video $x_0$ over $T$ discrete timesteps. This is a fixed Markov process defined as
$q(x_{1:T} | x_0) = \prod_{t=1}^T q(x_t | x_{t-1})$, 
where $q(x_t | x_{t-1}) = \mathcal{N}(x_t; \sqrt{1 - \beta_t} x_{t-1}, \beta_t \mathbf{I})$, and $\beta_t$ is a predefined noise schedule. And we can sample $x_t$ at any arbitrary timestep $t$ in a closed form:
$q(x_t | x_0) = \mathcal{N}(x_t; \sqrt{\bar{\alpha}_t} x_0, (1 - \bar{\alpha}_t) \mathbf{I})$, 
where $\alpha_t = 1 - \beta_t$ and $\bar{\alpha}_t = \prod_{i=1}^t \alpha_i$. As $t \to T$, $x_T$ approaches a pure Gaussian noise $\mathcal{N}(0, \mathbf{I})$.

\textbf{Reverse Process.}
The reverse process $p_\theta$ is a neural network trained to denoise $x_t$ back to $x_{t-1}$ for all $t \in [1, T]$. This process is parameterized as a Gaussian:
$p_\theta(x_{t-1} | x_t, c) = \mathcal{N}(x_{t-1}; \mu_\theta(x_t, t, c), \Sigma_\theta(x_t, t, c))$, 
where $c$ represents the conditioning information. 
In practice, the network $\epsilon_\theta$ is trained to predict the noise $\epsilon$ added to $x_0$ to create $x_t$. 
During inference, a video $x_0$ is generated by iteratively sampling $x_{t-1} \sim p_\theta(x_{t-1} | x_t, c)$ starting from pure noise $x_T \sim \mathcal{N}(0, \mathbf{I})$. This iterative process is the source of the train-inference discrepancy that our work addresses.

\subsection{3D Assets Harvesting}
\label{subsec:3dassets}

The rapid progress of 3D reconstruction has made it possible to lift 2D video frames back into 3D space, enabling game-engine–style manipulation while preserving real-world photorealism. Our framework begins by harvesting diverse 3D assets that support fine-grained scene editing with both geometric and appearance consistency. To systematically compare different 3D representations, we extract Gaussian splats, colored LiDAR point clouds, and 3D bounding boxes. Gaussian splats offer a dense, view-consistent radiance representation; colored LiDAR point clouds provide strong novel-view synthesis performance as demonstrated in StreetCrafter~\cite{yan2025streetcrafter}; and 3D bounding boxes remain a common control signal in generation methods~\cite{ren2025cosmos,gao2023magicdrive}.

We obtain high-quality Gaussian splats using OmniRe~\cite{chen2024omnire}, optimized across multiple nodes for large-scale driving logs. Following StreetCrafter~\cite{yan2025streetcrafter}, we harvest colored LiDAR point clouds by projecting 3D points into the image plane for color assignment. 3D bounding boxes are extracted directly from the Waymo dataset~\cite{sun2020waymo}. Our study in Section~\ref{ablation} shows that Gaussian splats preserve fine appearance details significantly better than sparse 3D primitives such as bounding boxes or colored point clouds.

However, although Gaussian splats excel at editing existing vehicles, inserting new assets remains challenging. To address this limitation, we incorporate mesh representations, which can be generated from text using off-the-shelf foundation models such as Hunyuan3D~\cite{zhao2025hunyuan3d}. We introduce two complementary pipelines to enable mesh-based insertion: \textit{3D Mesh Harvesting} and \textit{Inference-Time Insertion}. The first pipeline extracts meshes from video and reinserts them during training to reduce the photorealism gap between meshes and real imagery. The second pipeline converts arbitrary text descriptions into insertable 3D assets for downstream controllable editing tasks.

\begin{figure}[t]
  \centering
  \includegraphics[width=\linewidth]
  {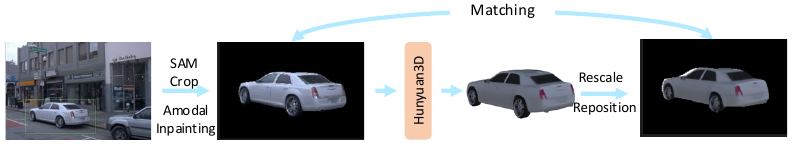}
  \vspace{-10pt}
  \caption{A demonstration of 3D Meshes Harvesting pipeline. 
  }
  \label{fig:mesh_pipeline}
  \vspace{-15pt}
\end{figure} 

\textbf{3D Meshes Harvesting.} 
As shown in Figure~\ref{fig:mesh_pipeline}, for mesh harvesting, we extract vehicle meshes from the Waymo dataset~\cite{sun2020waymo} using ground-truth 3D bounding boxes and LiDAR points. For each object, we first identify its best observation frame $f^*$, defined as the frame containing the largest number of associated LiDAR points, which typically provides minimal occlusion and the closest viewpoint.
We then apply SAM~\cite{kirillov2023segment-anything} to segment and crop the corresponding image, producing $C_{f^*}$. To handle occlusions, we perform Amodal inpainting using Pix2Gestalt~\cite{ozguroglu2024pix2gestalt}. The inpainted crop is then passed to Hunyuan3D~\cite{zhao2025hunyuan3d} to obtain a raw mesh:
\begin{equation}
M_{\text{raw}} = \mathcal{H}(C_{f^*}).
\end{equation}

Meshes produced by Hunyuan3D do not preserve the vehicle’s original pose or metric dimensions. Since accurate geometry is required for re-insertion during rendering, we need to align each generated mesh with its ground-truth 3D bounding box. While vehicles typically have a dominant length dimension, and Hunyuan3D outputs a consistent upright orientation, reducing the candidate poses to two. We then further use a GPT-based heading reasoning module~\cite{achiam2023gpt} to select the correct orientation. 

Given the correct rotation, we assume the mesh center coincides with the center of the ground-truth box and optimize only the scale $s$. We jointly minimize a depth discrepancy term and an IoU-based alignment term:
\begin{equation}
\text{Score} = \| D(M, s) - D_{\text{gt}} \| - \lambda \, \text{IoU}(B(M, s), B_{\text{gt}}),
\end{equation}
\begin{equation}
s^* = \arg\min_s \, \text{Score},
\end{equation}
where $D(\cdot)$ denotes rendered depth and $B(\cdot)$ denotes the projected bounding box. We solve for $s^*$ using a grid search initialized around the ground-truth box dimensions.

\textbf{Inference-Time Insertion.}
At inference, to enable inserting any vehicles with a text description $P$, we start from leveraging GPT~\cite{achiam2023gpt} to produce a reference image $I_{\text{ref}}$ and feed it into Hunyuan3D~\cite{zhao2025hunyuan3d} to obtain $M_{\text{gen}}$. 
Afterwards, the rotation and scale would be determined with a Vision Language Model reasoning process leveraging GPT~\cite{achiam2023gpt}.  
This guarantees that all inserted or modified vehicles follow their physically valid paths without spatial conflict. 

\begin{figure}[!htbp]
  \centering
  \includegraphics[width=.9\linewidth]
  {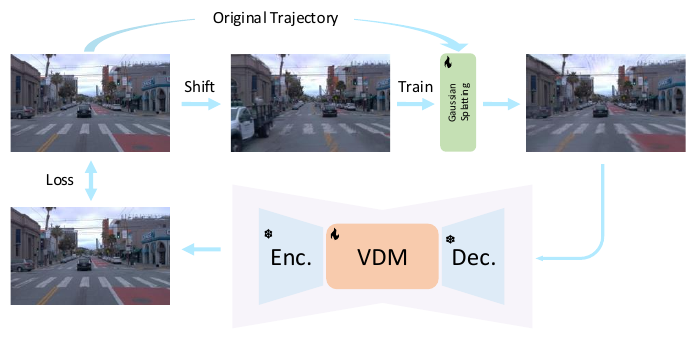}
  \vspace{-5pt}
  \caption{A demonstration of constructing data pairs for Gaussian Splats. 
  }
  \label{fig:training_pipeline}
  \vspace{-15pt}
\end{figure}

\subsection{Video Rendering}
\label{subsec:video_rendering}

With the harvested 3D assets, we can reposition and manipulate any agent in the scene according to a target trajectory. However, rendering these 3D assets back into 2D videos inevitably introduces artifacts, as perfect 3D reconstruction is difficult, if not impossible, to obtain. To bridge this gap, we introduce a video rendering module that maps imperfect 3D renderings into photorealistic video sequences. This section first describes how we construct training pairs for each representation, followed by the diffusion-based training objective.

\textbf{Constructing data pairs for Gaussian Splats.}
Artifacts in Gaussian splats commonly arise when rendering from novel viewpoints. A robust rendering module must therefore learn to correct such distortions and synthesize high-quality, temporally consistent video. Motivated by prior work on reconstruction-based denoising~\cite{ni2025recondreamer,wu2025difix3d}, we adopt a cycle-reconstruction strategy similar to~\cite{wu2025difix3d} as shown in Figure~\ref{fig:training_pipeline}.

Given a Gaussian field $G_1$ reconstructed from a real scene under trajectories $\mathcal{T}$, we perturb the trajectories to obtain $\tilde{\mathcal{T}}$:
\begin{equation}
\tilde{\mathcal{T}} = \mathcal{T} + \Delta \mathcal{T}, \qquad 
\Delta \mathcal{T} = \{\delta_i\}_{i=1}^N,
\end{equation}
where each $\delta_i$ applies horizontal or vertical shifts or a change in heading angle.
We then render perturbed frames $\tilde{v}$ from $G_1$ under $\tilde{\mathcal{T}}$, train a new Gaussian field $G_2$ on $\tilde{v}$, and render $G_2$ back under the original $\mathcal{T}$ to form clean–noisy training pairs:
\begin{equation}
G_2 = \mathcal{T}_{\text{recon}}(\tilde{v}), \qquad 
\hat{v} = \mathcal{R}(G_2, \mathcal{T}).
\end{equation}
This process yields paired samples $(\tilde{v}, v)$ that enable the rendering module to learn to remove Gaussian-splat artifacts and restore photorealistic appearance.

\textbf{Constructing data pairs for Meshes.}
Meshes often exhibit a substantial photorealism gap compared to real driving videos. To train a rendering model that can compensate for this discrepancy, we construct paired training examples by randomly replacing Gaussian-splat assets with harvested meshes during rendering. For each scene, we synthesize two versions: one with mesh-insertion probability $p = 0.5$ and one with $p = 0.0$ (mimicking scenes without object insertion). We additionally randomize Gaussian-splat lighting conditions to enhance robustness to illumination differences between mesh renderings and real scenes.

\begin{figure}[!htbp]
\vspace{-5pt}
  \centering
  \includegraphics[width=.8\linewidth]
  {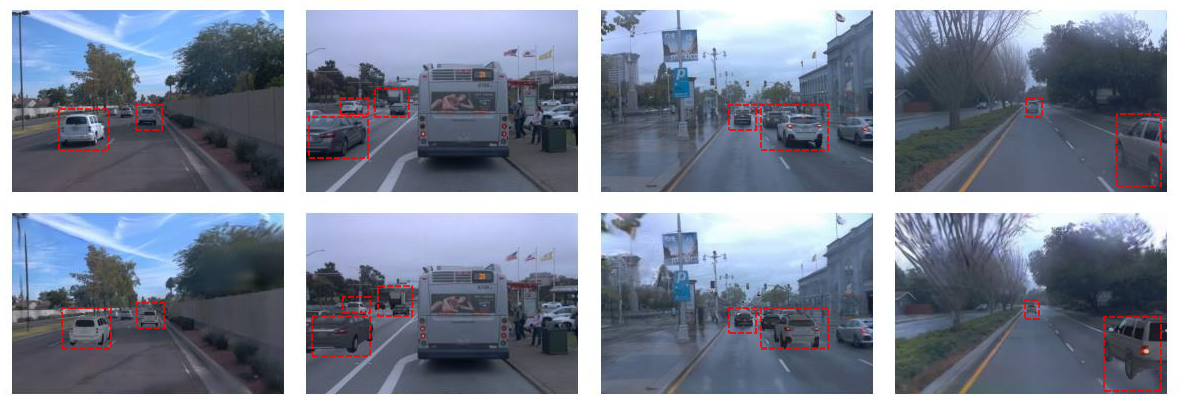}
  \vspace{-5pt}
  \caption{A demonstration of mesh-Gaussian training data pairs. The top frames are the original Gaussian Splats and the bottom ones are Gaussian with mesh vehicle replacements}
  \label{fig:pair_demo}
  \vspace{-5pt}
\end{figure}

For rendering compatibility across representations, Gaussian RGB and depth maps are rendered using OmniRe~\cite{chen2024omnire}, while mesh RGB and depth are generated using PyVista~\cite{sullivan2019pyvista}. The two modalities are then merged via depth composition. Example data pairs are shown in Figure~\ref{fig:pair_demo}.

\textbf{Diffusion Objective.}
For each training pair, we render the conditioning video $v_c$ from Gaussian splats or meshes under synchronized camera poses and train a video diffusion model to map these imperfect renderings to photorealistic outputs. 
Given a clean target video $x_0$ and a timestep $t \sim \mathcal{U}\{1,\ldots,T\}$, we sample noise $\epsilon \sim \mathcal{N}(0,\mathbf{I})$ and construct the noisy input
\begin{equation}
x_t = \sqrt{\bar{\alpha}_t}\, x_0 \;+\; \sqrt{1-\bar{\alpha}_t}\, \epsilon.
\end{equation}
The model predicts $\epsilon_\theta(x_t, t, v_c)$ and is trained with the standard denoising objective
\begin{equation}
\mathcal{L}_{\text{vdm}}
  = \mathbb{E}_{t,\epsilon}\Big[\,\|\epsilon - \epsilon_\theta(x_t, t, v_c)\|_2^2\,\Big].
\end{equation}
This formulation works effectively when paired with explicit 3D conditioning, resulting in stable and coherent multi-agent video synthesis. In practice, we fine-tune the TrajectoryCrafter pretrained model~\cite{mark2025trajectorycrafter} using the CogVideoX backbone~\cite{yang2024cogvideox} for video diffusion.

We additionally compare this video-based rendering module with an image diffusion approach following~\cite{parmar2024one,wu2025difix3d}. As shown in Section~\ref{ablation}, our experiments show that video diffusion yields substantially better results, benefiting from richer temporal priors and improved cross-frame consistency.

\subsection{\ourbenchmark}
\label{subsec:benchmark}

To quantitatively evaluate controllable multi-agent scene generation, we introduce \textbf{\ourbenchmark}, a comprehensive benchmark that assesses both ego-vehicle and surrounding-vehicle manipulations under a unified trajectory set $\mathcal{T}$.

\begin{figure}[!htbp]
\vspace{-10pt}
  \centering
  \includegraphics[width=.8\linewidth]
  {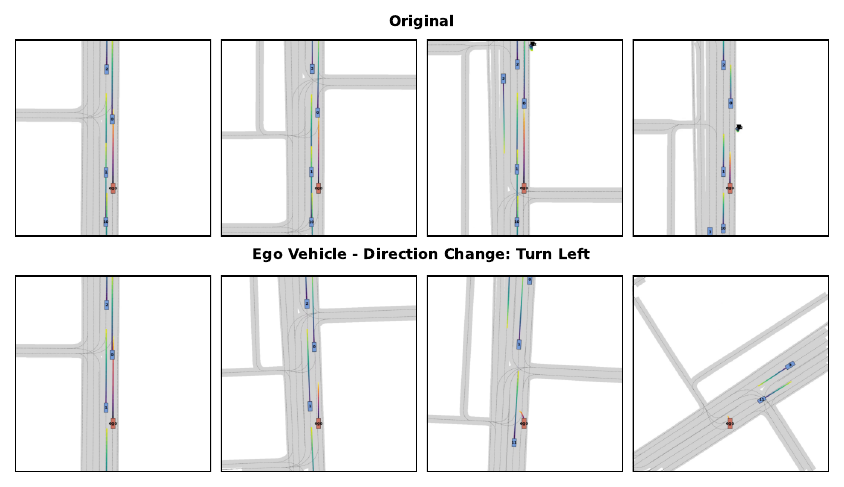}
  \vspace{-5pt}
  \caption{A demonstration for trajectory modification in our proposed \textbf{\ourbenchmark}. }
  \label{fig:benchmark_demo}
  \vspace{-10pt}
\end{figure}

Unlike prior benchmarks that primarily focus on ego lane-changing performance~\cite{ni2025recondreamer,yan2025streetcrafter}, \textbf{\ourbenchmark} targets realistic editing requirements encountered in real-world driving scenarios. We define a diverse suite of manipulation tasks for both ego and non-ego agents, including speed changes, lane changes, direction changes, insertions, and removals. For each task category, we begin by describing the desired trajectory modification in natural language, then apply the language-driven trajectory editing model LangTraj~\cite{chang2025langtraj} to produce corresponding trajectory edits within the scenes sampled from Waymo evaluation dataset~\cite{sun2020waymo}. Afterward, we manually select 109 challenging, high-quality edited trajectories spanning all manipulation types, forming a rigorous testbed for controllable trajectory editing. Figure~\ref{fig:benchmark_demo} illustrates an example of ego-vehicle direction change. Details of \textbf{\ourbenchmark} are provided in Appendix~\ref{appendix:benchmark}.

\begin{table*}[t]
\centering
\resizebox{.85\linewidth}{!}{
\begin{tabular}{lcccccccccccc}
\toprule
\multicolumn{13}{c}{\textbf{(a) Ego Car Manipulation}}\\
\midrule
\multirow{2}{*}{Method} & 
\multicolumn{4}{c}{Speed Change} & 
\multicolumn{4}{c}{Lane Change} & 
\multicolumn{4}{c}{Direction Change} \\
\cmidrule(lr){2-5} \cmidrule(lr){6-9} \cmidrule(lr){10-13}
 & FID$\downarrow$ & FVD$\downarrow$ & VIMS$\uparrow$ & BAS$\uparrow$ 
 & FID$\downarrow$ & FVD$\downarrow$ & VIMS$\uparrow$ & BAS$\uparrow$
 & FID$\downarrow$ & FVD$\downarrow$ & VIMS$\uparrow$ & BAS$\uparrow$ \\
\midrule
StreetCrafter~\cite{yan2025streetcrafter} & 150.11 & 1232.13 & 0.9623 & 0.8754 & 134.65 & 1347.02 & 0.9568 & 0.8612 & 199.28 & 1846.96 & 0.9573 & 0.8497 \\
Difix3D~\cite{wu2025difix3d} & 146.11 & 1019.25 & \cellcolor{green!5}\underline{0.9643} & 0.8357 & 135.80 & 1049.53 & \cellcolor{green!5}\underline{0.9586} & 0.8191 & 203.11 & 1416.44 & \cellcolor{green!15}\textbf{0.9590} & 0.8052 \\
OmniRe~\cite{chen2024omnire} & \cellcolor{green!5}\underline{46.89} & \cellcolor{green!15}\textbf{548.22} & 0.9619 & \cellcolor{green!5}\underline{0.9121} & \cellcolor{green!5}\underline{54.02} & \cellcolor{green!15}\textbf{667.70} & 0.9541 & \cellcolor{green!5}\underline{0.9102} & \cellcolor{green!15}\textbf{80.60} & \cellcolor{green!5}\underline{1237.23} & 0.9517 & \cellcolor{green!5}\underline{0.8719} \\
\textbf{Ours} & \cellcolor{green!15}\textbf{39.92} & \cellcolor{green!5}\underline{583.78} & \cellcolor{green!15}\textbf{0.9657} & \cellcolor{green!15}\textbf{0.9326} & \cellcolor{green!15}\textbf{50.33} & \cellcolor{green!5}\underline{729.76} & \cellcolor{green!15}\textbf{0.9615} & \cellcolor{green!15}\textbf{0.9303} & \cellcolor{green!5}\underline{144.83} & \cellcolor{green!15}\textbf{1175.52} & \cellcolor{green!5}\underline{0.9585} & \cellcolor{green!15}\textbf{0.8947}\\
\midrule
\multicolumn{13}{c}{\textbf{(b) Other Vehicle Manipulation I}}\\
\midrule
\multirow{2}{*}{Method} & 
\multicolumn{4}{c}{Speed Change} & 
\multicolumn{4}{c}{Lane Change} & 
\multicolumn{4}{c}{Direction Change} \\
 \cmidrule(lr){2-5} \cmidrule(lr){6-9} \cmidrule(lr){10-13}
 & FID$\downarrow$ & FVD$\downarrow$ & VIMS$\uparrow$ & BAS$\uparrow$ 
 & FID$\downarrow$ & FVD$\downarrow$ & VIMS$\uparrow$ & BAS$\uparrow$
 & FID$\downarrow$ & FVD$\downarrow$ & VIMS$\uparrow$ & BAS$\uparrow$ \\
\midrule
StreetCrafter~\cite{yan2025streetcrafter} & 102.90 & 1181.27 & 0.9674 & 0.8887 & 163.21 & 1467.54 & 0.9600 & 0.8818 & 140.32 & 1310.09 & 0.9683 & 0.8820\\
Difix3D~\cite{wu2025difix3d}  & 128.27 & 1038.93 & \cellcolor{green!5}\underline{0.9716} & 0.8500 & 149.11 & 1358.60 & \cellcolor{green!5}\underline{0.9624} & 0.8449 & 133.91 & 994.03 & \cellcolor{green!5}\underline{0.9708} & 0.8387\\
OmniRe~\cite{chen2024omnire} & \cellcolor{green!5}\underline{38.83} & \cellcolor{green!5}\underline{527.26} & 0.9711 & \cellcolor{green!5}\underline{0.9355} & \cellcolor{green!5}\underline{48.05} & \cellcolor{green!5}\underline{580.17} & 0.9619 & \cellcolor{green!5}\underline{0.9238} & \cellcolor{green!5}\underline{70.35} & \cellcolor{green!5}\underline{598.01} & 0.9707 & \cellcolor{green!5}\underline{0.9159}\\
\textbf{Ours} & \cellcolor{green!15}\textbf{38.62} & \cellcolor{green!15}\textbf{499.43} & \cellcolor{green!15}\textbf{0.9770} & \cellcolor{green!15}\textbf{0.9512} & \cellcolor{green!15}\textbf{44.34} & \cellcolor{green!15}\textbf{539.40} & \cellcolor{green!15}\textbf{0.9633} & \cellcolor{green!15}\textbf{0.9391} & \cellcolor{green!15}\textbf{54.54} & \cellcolor{green!15}\textbf{508.53} & \cellcolor{green!15}\textbf{0.9732} & \cellcolor{green!15}\textbf{0.9333}\\
\midrule
\multicolumn{13}{c}{\textbf{(c) Other Vehicle Manipulation II \& Overall Assessment}}\\
\midrule
\multirow{2}{*}{Method} & 
\multicolumn{5}{c}{Insertion} & 
\multicolumn{5}{c}{Removal}  &
\multicolumn{2}{c}{Overall}
 \\
\cmidrule(lr){2-6} \cmidrule(lr){7-11} \cmidrule(lr){12-13} 
 & FID$\downarrow$ & FVD$\downarrow$ & VIMS$\uparrow$ & BAS$\uparrow$  & OSR$\uparrow$  
 & FID$\downarrow$ & FVD$\downarrow$ & VIMS$\uparrow$ & BAS$\uparrow$  & OSR$\uparrow$ 
 & FID$\downarrow$ & FVD$\downarrow$
\\
\midrule
StreetCrafter~\cite{yan2025streetcrafter} & 267.76 & 2558.27 & 0.9652 & 0.8276 & 4.62 & 92.19 & 1209.77 & 0.9785 & 0.8918 & 6.51 & 91.16 & 1245.96  \\
Difix3D~\cite{wu2025difix3d} & 267.53 & 2122.44 & 0.9664 & 0.7824 & \cellcolor{green!5}\underline{4.71} & 122.80 & 786.16 & \cellcolor{green!5}\underline{0.9823} & 0.8526 & \cellcolor{green!5}\underline{6.89} & 80.84 & 991.23\\
OmniRe~\cite{chen2024omnire} & \cellcolor{green!5}\underline{182.29} & \cellcolor{green!5}\underline{1636.95} & \cellcolor{green!5}\underline{0.9754} & \cellcolor{green!5}\underline{0.8732} & 4.23 & \cellcolor{green!5}\underline{33.97} & \cellcolor{green!5}\underline{390.97} & 0.9712 & \cellcolor{green!5}\underline{0.9363} & 6.42 & \cellcolor{green!5}\underline{44.37} & \cellcolor{green!5}\underline{546.00}\\
\textbf{Ours} & \cellcolor{green!15}\textbf{117.46} & \cellcolor{green!15}\textbf{1142.09} & \cellcolor{green!15}\textbf{0.9784} & \cellcolor{green!15}\textbf{0.8913} & \cellcolor{green!15}\textbf{5.86} & \cellcolor{green!15}\textbf{29.01} & \cellcolor{green!15}\textbf{380.71} & \cellcolor{green!15}\textbf{0.9884} & \cellcolor{green!15}\textbf{0.9579} & \cellcolor{green!15}\textbf{7.00} & \cellcolor{green!15}\textbf{33.19} & \cellcolor{green!15}\textbf{536.49}\\
\bottomrule
\end{tabular}}
\caption{Quantitative comparison on \textbf{\ourbenchmark} with different methods. 
Metrics include FID$\downarrow$, FVD$\downarrow$, Vehicle Identity Matching Score (VIMS)$\uparrow$, Background Alignment Score (BAS)$\uparrow$ and Operation Success Rate (OSR)$\uparrow$. 
Lower is better for $\downarrow$ and higher is better for $\uparrow$. 
The best results are highlighted in \colorbox{green!15}{\textbf{bold}}, and the second best results are \colorbox{green!5}{\underline{underlined}}.}
\label{tab:quant_comp}
\vspace{-10pt}
\end{table*}

\begin{figure*}[t]
  \centering
  \includegraphics[width=.8\textwidth]{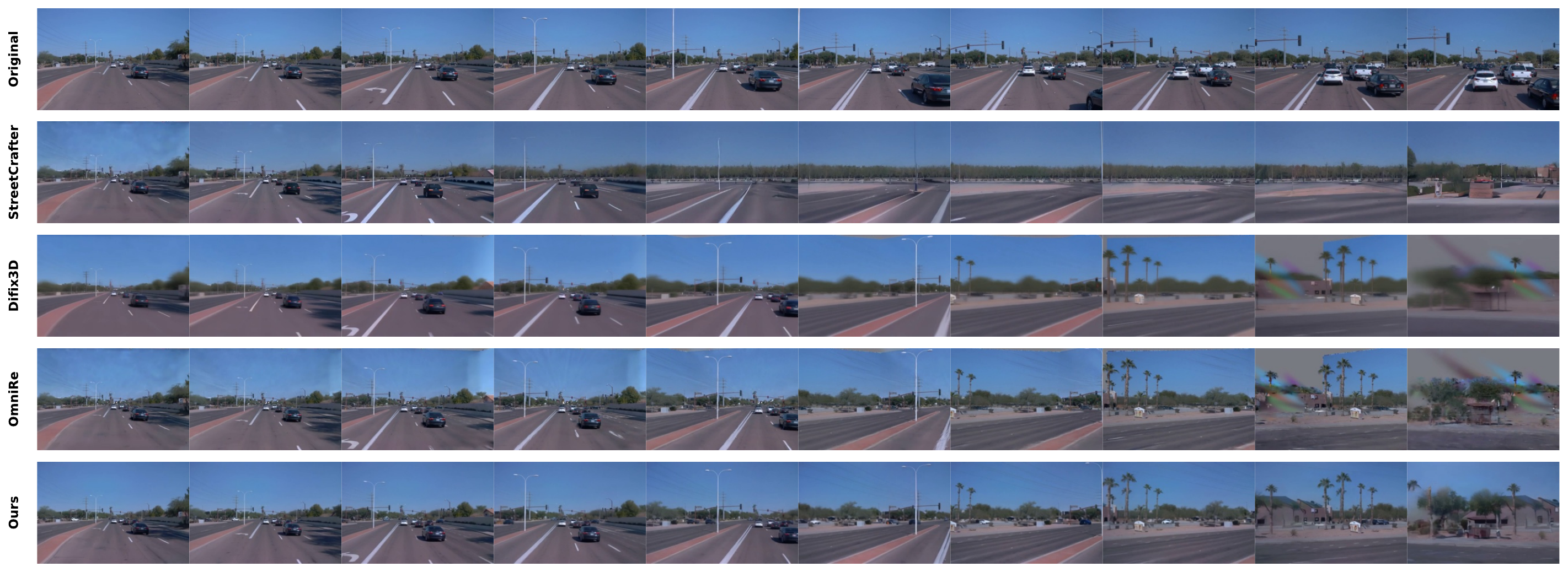}
  \caption{Qualitative Comparison among different methods. From top to bottom is Original scene, StreetCrafter~\cite{yan2025streetcrafter}, Difix3D~\cite{wu2025difix3d}, OmniRe~\cite{chen2024omnire} and \textbf{Ours}. It's a operation of turning ego direction to left. 
  }
  \label{fig:quality_comp}
  \vspace{-15pt}
\end{figure*}

To rigorously assess video quality, temporal consistency, and alignment with both the source videos and input trajectories, we employ the following five metrics:
\begin{itemize}
\item \textbf{FID \& FVD}~\cite{heusel2017gans, unterthiner2018towards}: Fréchet Inception Distance and Fréchet Video Distance evaluate spatial fidelity and temporal realism of generated videos relative to the source sequences.
\item \textbf{Vehicle Identity Matching Score (VIMS)}: Based on the intuition that vehicles occupying similar positions in the ego coordinate frame should retain consistent appearance across the source and edited videos, VIMS jointly measures trajectory adherence and visual identity preservation. For each non-occluded vehicle in each frame, we locate its source counterpart with a similar ego-frame pose and compute their CLIP similarity. The final score averages these similarities across all vehicles and frames. A formal definition is provided in Appendix~\ref{appendix:vims}. 
\item \textbf{Background Alignment Score (BAS)}: BAS quantifies static-scene preservation by computing CLIP similarity~\cite{radford2021learning} between the background regions of the edited frames and its closest source frames under ego coordinates after masking out all dynamic objects. A formal definition is provided in Appendix~\ref{appendix:bas}.
\item \textbf{Operation Success Rate (OSR)}: For insertion and removal tasks, we measure semantic correctness using GPT-based reasoning~\cite{achiam2023gpt}. The model assigns a score from 1 to 10 reflecting whether the manipulation is correctly realized and visually realistic. A formal definition is provided in Appendix~\ref{appendix:osr}.
\end{itemize}
Together, \textbf{\ourmethod} and \textbf{\ourbenchmark} establish a unified foundation for multi-agent, physically consistent, and photorealistic driving-scene editing and evaluation.

\vspace{-5pt}
\section{Experiments}

In this section, we conduct a series of experiments to validate the effectiveness of our proposed \textbf{\ourmethod} comparing with multiple baselines and settings on \textbf{\ourbenchmark}. We finetuned our model for 60k steps based on CogVideoX~\cite{yang2024cogvideox} backbone, utilizing AdamW~\cite{loshchilov2017decoupled} as optimizer. For more implementation details, please refer to Appendix~\ref{appendix:imp}.
In Section~\ref{qual_comp}, we evaluate its task performances comparing against state-of-the-art and representative rendering methods to demonstrate its efficacy. 
In Section~\ref{ablation}, we present systematic ablation studies to explore the impact of different modalities on autonomous driving scene rendering tasks and the importance of temporal consistency for such tasks. 
Finally, Section~\ref{user_study} further conducts a comprehensive user study to demonstrate that the quality of videos generated by \textbf{\ourmethod} are aligned with human preference. As for downstream task, please refer to Appendix~\ref{appendix:down}. 

\subsection{Quality Comparisons}\label{qual_comp}
We compare \textbf{\ourmethod} with three representative baselines covering different paradigms of controllable scene generation: 
StreetCrafter~\cite{yan2025streetcrafter}, Difix3D~\cite{wu2025difix3d} and 
OmniRe~\cite{chen2024omnire}
. 
All methods are evaluated on \textbf{\ourbenchmark} under the same settings, covering both ego-vehicle and multi-agent manipulation tasks such as speed change, lane change, direction change, insertion, and removal. 
As shown in Table~\ref{tab:quant_comp} and Figure~\ref{fig:quality_comp}, \textbf{\ourmethod} consistently outperforms all baselines across almost all metrics and achieves the best overall FID and FVD scores. 
Compared with baselines, 
our Gaussian–mesh representation knowledge demonstrates superior generalizability and produces finer appearance details and preserves stable geometry under strong ego and agent motion. 
These results showcase that \textbf{\ourmethod}, integrating editable Gaussian splats and meshes with a video diffusion backbone, provides a simple yet powerful paradigm for photorealistic, physically reliable, and controllable driving scene generation.

\begin{table*}[t] 
\centering
\resizebox{.85\linewidth}{!}{
\begin{tabular}{lcccccccccccc}
\toprule
\multicolumn{13}{c}{\textbf{(a) Ego Car Manipulation}}\\
\midrule
\multirow{2}{*}{Method} & 
\multicolumn{4}{c}{Speed Change} & 
\multicolumn{4}{c}{Lane Change} & 
\multicolumn{4}{c}{Direction Change} \\
\cmidrule(lr){2-5} \cmidrule(lr){6-9} \cmidrule(lr){10-13}
 & FID$\downarrow$ & FVD$\downarrow$ & VIMS$\uparrow$ & BAS$\uparrow$ 
 & FID$\downarrow$ & FVD$\downarrow$ & VIMS$\uparrow$ & BAS$\uparrow$
 & FID$\downarrow$ & FVD$\downarrow$ & VIMS$\uparrow$ & BAS$\uparrow$ \\
\midrule
Point Cloud  & \cellcolor{green!5}\underline{62.16} & \cellcolor{green!5}\underline{869.06} & 0.9633 & \cellcolor{green!5}\underline{0.9051} & \cellcolor{green!5}\underline{81.65} & \cellcolor{green!5}\underline{877.87} & 0.9599 & \cellcolor{green!5}\underline{0.8946} & 152.10 & 1427.25 & 0.9593 & \cellcolor{green!5}\underline{0.8751}\\
BBox  & 108.54 & 2105.98 & 0.9491 & 0.7986 & 125.19 & 1677.12 & 0.9469 & 0.7744 & 228.06 & 2662.92 & 0.9475 & 0.7663\\
\cmidrule(lr){1-1}\cmidrule(lr){2-13}
Image DM & 82.19 & 920.29 & \cellcolor{green!5}\underline{0.9655} & 0.8353 & 89.77 & 967.54 & \cellcolor{green!5}\underline{0.9607} & 0.8188 & \cellcolor{green!15}\textbf{114.11} & \cellcolor{green!5}\underline{1180.18} & \cellcolor{green!15}\textbf{0.9608} & 0.8063\\
\cmidrule(lr){1-1}\cmidrule(lr){2-13}
\textbf{Ours} & \cellcolor{green!15}\textbf{39.92} & \cellcolor{green!15}\textbf{583.78} & \cellcolor{green!15}\textbf{0.9657} & \cellcolor{green!15}\textbf{0.9326} & \cellcolor{green!15}\textbf{50.33} & \cellcolor{green!15}\textbf{729.76} & \cellcolor{green!15}\textbf{0.9615} & \cellcolor{green!15}\textbf{0.9303} & \cellcolor{green!5}\underline{144.83} & \cellcolor{green!15}\textbf{1175.52} & \cellcolor{green!5}\underline{0.9595} & \cellcolor{green!15}\textbf{0.8947}\\
\midrule
\multicolumn{13}{c}{\textbf{(b) Other Vehicle Manipulation I}}\\
\midrule
\multirow{2}{*}{Method} & 
\multicolumn{4}{c}{Speed Change} & 
\multicolumn{4}{c}{Lane Change} & 
\multicolumn{4}{c}{Direction Change} \\
 \cmidrule(lr){2-5} \cmidrule(lr){6-9} \cmidrule(lr){10-13}
 & FID$\downarrow$ & FVD$\downarrow$ & VIMS$\uparrow$ & BAS$\uparrow$ 
 & FID$\downarrow$ & FVD$\downarrow$ & VIMS$\uparrow$ & BAS$\uparrow$
 & FID$\downarrow$ & FVD$\downarrow$ & VIMS$\uparrow$ & BAS$\uparrow$ \\
\midrule
Point Cloud  & \cellcolor{green!5}\underline{69.47} & \cellcolor{green!5}\underline{844.01} & 0.9714 & \cellcolor{green!5}\underline{0.9077} & \cellcolor{green!5}\underline{74.80} & \cellcolor{green!5}\underline{905.17} & 0.9598 & \cellcolor{green!5}\underline{0.9023} & \cellcolor{green!5}\underline{84.89} & 840.43 & 0.9699 & \cellcolor{green!5}\underline{0.8975}\\
BBox  & 100.61 & 1428.47 & 0.9584 & 0.8156 & 122.76 & 1994.33 & 0.9501 & 0.8001 & 124.87 & 1499.41 & 0.9587 & 0.8304\\
\cmidrule(lr){1-1}\cmidrule(lr){2-13}
Image DM & 88.99 & 870.43 & \cellcolor{green!5}\underline{0.9733} & 0.8416 & 89.44 & 1271.34 & \cellcolor{green!15}\textbf{0.9636} & 0.8470 & 103.89 & \cellcolor{green!5}\underline{821.33} & \cellcolor{green!5}\underline{0.9720} & 0.8386\\
\cmidrule(lr){1-1}\cmidrule(lr){2-13}
\textbf{Ours} & \cellcolor{green!15}\textbf{38.62} & \cellcolor{green!15}\textbf{499.43} & \cellcolor{green!15}\textbf{0.9770} & \cellcolor{green!15}\textbf{0.9512} & \cellcolor{green!15}\textbf{44.34} & \cellcolor{green!15}\textbf{539.40} & \cellcolor{green!5}\underline{0.9633} & \cellcolor{green!15}\textbf{0.9391} & \cellcolor{green!15}\textbf{54.54} & \cellcolor{green!15}\textbf{508.53} & \cellcolor{green!15}\textbf{0.9732} & \cellcolor{green!15}\textbf{0.9333}\\
\midrule
\multicolumn{13}{c}{\textbf{(c) Other Vehicle Manipulation II \& Overall Assessment}}\\
\midrule
\multirow{2}{*}{Method} & 
\multicolumn{5}{c}{Insertion} & 
\multicolumn{5}{c}{Removal}  &
\multicolumn{2}{c}{Overall}
 \\
\cmidrule(lr){2-6} \cmidrule(lr){7-11} \cmidrule(lr){12-13} 
  & FID$\downarrow$ & FVD$\downarrow$ & VIMS$\uparrow$ & BAS$\uparrow$  & OSR$\uparrow$  
 & FID$\downarrow$ & FVD$\downarrow$ & VIMS$\uparrow$ & BAS$\uparrow$  & OSR$\uparrow$ 
 & FID$\downarrow$ & FVD$\downarrow$
\\
\midrule
Point Cloud  & \cellcolor{green!5}\underline{148.92} & 1665.92 & 0.9713 & 0.8633 & \cellcolor{green!15}\textbf{6.07} & \cellcolor{green!5}\underline{56.55} & 763.86 & 0.9807 & \cellcolor{green!5}\underline{0.9139} & 6.68 & \cellcolor{green!5}\underline{54.14} & \cellcolor{green!5}\underline{813.67}\\
BBox  & 166.21 & 2645.56 & 0.9621 & 0.7707 & 4.28 & 87.12 & 1351.49 & 0.9672 & 0.8203 & 6.89 & 81.74 & 1521.07\\
\cmidrule(lr){1-1}\cmidrule(lr){2-13}
Image DM & 176.96 & \cellcolor{green!5}\underline{1615.10} & \cellcolor{green!5}\underline{0.9758} & \cellcolor{green!15}\textbf{0.8995} & 4.00 & 74.29 & \cellcolor{green!5}\underline{703.47} & \cellcolor{green!5}\underline{0.9842} & 0.8454 & \cellcolor{green!15}\textbf{7.00} & 75.83 & 837.99\\
\cmidrule(lr){1-1}\cmidrule(lr){2-13}
\textbf{Ours} & \cellcolor{green!15}\textbf{117.46} & \cellcolor{green!15}\textbf{1142.09} & \cellcolor{green!15}\textbf{0.9784} & \cellcolor{green!5}\underline{0.8913} & \cellcolor{green!5}\underline{5.86} & \cellcolor{green!15}\textbf{29.01} & \cellcolor{green!15}\textbf{380.71} & \cellcolor{green!15}\textbf{0.9884} & \cellcolor{green!15}\textbf{0.9579} & \cellcolor{green!15}\textbf{7.00} & \cellcolor{green!15}\textbf{33.19} & \cellcolor{green!15}\textbf{536.49}\\
\bottomrule
\end{tabular}}

\caption{Ablation study on \textbf{\ourbenchmark} with different settings. 
Metrics include FID$\downarrow$, FVD$\downarrow$, Vehicle Identity Matching Score (VIMS)$\uparrow$, Background Alignment Score (BAS)$\uparrow$ and Operation Success Rate (OSR)$\uparrow$. 
Lower is better for $\downarrow$ and higher is better for $\uparrow$. 
The best results are highlighted in \colorbox{green!15}{\textbf{bold}}, and the second best results are \colorbox{green!5}{\underline{underlined}}.}
\label{tab:ablation}
\vspace{-15pt}
\end{table*}

\vspace{-2pt}
\subsection{Ablation Study}\label{ablation}
\vspace{-3pt}

We conduct detailed ablations on \textbf{\ourmethod} using \textbf{\ourbenchmark} to analyze the contribution of different conditioning modalities and model designs. Specifically, we compare (1) different 3D conditioning inputs for the diffusion model and (2) the effectiveness of using a video diffusion model (VDM) versus an image diffusion model (Image DM). The quantitative results are reported in Table~\ref{tab:ablation}.

For the first aspect, we compare our Gaussian–mesh conditioning with alternative 3D modalities, including colored point clouds~\cite{yan2025streetcrafter} and bounding boxes (BBoxes)~\cite{sun2020waymo}. For fair comparison, we combine all of them with meshes. The results show that Gaussian representation provides competitive geometric and photometric knowledge for high-quality, controllable scene generation. While colored point clouds capture accurate spatial positions and exhibit robustness to trajectory perturbations, their sparsity and lack of fine appearance cues limit their ability to guide diffusion synthesis at high fidelity. BBoxes, on the other hand, offer precise localization and motion information but fail to encode structural or visual context, resulting in unstable textures and poor alignment with the original scene. 
In contrast, Gaussian splats inherently encode both continuous density and color information, providing rich mid-level priors that effectively balance spatial precision and appearance detail, leading to the overall best performances across all manipulation categories.
For the second aspect, we compare our full video diffusion model with a single-frame Image DM. We utilize Difix3D~\cite{wu2025difix3d} as backbone and finetune it under the same conditioning setup with our method. Although the Image DM baseline achieves competitive fidelity, it lacks temporal awareness and thus struggles to maintain cross-frame consistency. This leads to noticeable temporal flickering. In contrast, our \textbf{\ourmethod} explicitly models temporal correlations, preserving coherent motion across frames and maintaining consistent visual features during complex multi-agent edits. These findings highlight that temporal priors are indispensable for realistic controllable video generation, and that combining Gaussian–mesh conditioning with a temporal diffusion backbone yields the most stable and photorealistic results.

\subsection{User Study}\label{user_study}

\begin{table}[!htbp]
\centering
\resizebox{.8\linewidth}{!}{
\begin{tabular}{lccc}
\toprule
\textbf{Method} & \textbf{Wins} & \textbf{Total} & \textbf{Win Rate} \\
\midrule
StreetCrafter~\cite{yan2025streetcrafter} & 2 & 501 & 0.40\% \\
Difix3D~\cite{wu2025difix3d} & 4 & 501 & 0.80\% \\
OmniRe~\cite{chen2024omnire} & 39 & 501 & 7.78\% \\
\textbf{Ours} & \cellcolor{green!15}\textbf{456} & 501 & \cellcolor{green!15}\textbf{91.02\%} \\
\bottomrule
\end{tabular}}
\vspace{-5pt}
\caption{Comparison of win rates among different methods.}
\label{tab:user_study}
\vspace{-5pt}
\end{table}

To further assess the perceptual quality and controllability of generated videos, we conducted a comprehensive user study comparing under the same comparison setting shown in Sec.~\ref{qual_comp}. The study was designed to evaluate how well each method produces photorealistic and trajectory-consistent driving videos under identical editing tasks.
Each participant was shown 20 video samples randomly drawn from \textbf{\ourbenchmark}. For the design details, please refer to Appendix~\ref{appendix:user}.
As shown in Table~\ref{tab:user_study}, \textbf{\ourmethod} overwhelmingly outperformed other methods. These results strongly confirm that users consistently perceive \textbf{\ourmethod}’s outputs as more realistic, visually stable, and physically consistent with the intended trajectories, validating both its photorealism and controllability advantages.

\section{Conclusion}
This paper introduced \textbf{\ourmethod}, a simple yet effective framework for controllable driving scene generation that combines editable 3D Gaussian splats and meshes with a standard video diffusion model, enabling photorealistic and physically grounded multi-agent editing through trajectory-based control. \textbf{\ourmethod} supports fine-grained manipulation of both ego and surrounding vehicles and consistently outperforms reconstruction-based, generation-based, and hybrid baselines across all tasks in our newly proposed benchmark \textbf{\ourbenchmark}. Our systematic study further highlights the effectiveness of Gaussian conditioning and the importance of temporal diffusion for stable, coherent scene synthesis. Together, \textbf{\ourmethod} and \textbf{\ourbenchmark} provide a unified, scalable, and physically reliable foundation for future research in controllable driving simulation.
{
    \small
    \bibliographystyle{ieeenat_fullname}
    \bibliography{main}
}

\clearpage
\setcounter{page}{1}
\maketitlesupplementary

\section{Implementation Detail}\label{appendix:imp}

We fine-tune our Video Diffusion Model (VDM) using the 5B CogVideoX backbone~\cite{yang2024cogvideox}, initialized from the TrajectoryCrafter checkpoint~\cite{yu2025trajectorycrafter}. All experiments are implemented with \texttt{pytorch}~\cite{paszke2019pytorch}, \texttt{accelerate}~\cite{accelerate} and \texttt{diffuser}~\cite{von-platen-2022-diffusers}. Four H100~(80GB) GPUs are employed for training. It uses a per-GPU batch size of 1 with gradient accumulation of 1, yielding an effective batch size of 4. We adopt \texttt{bfloat16} mixed-precision training to reduce memory consumption while maintaining numerical stability.

We perform finetuning in two stages. In the first stage, we train for 40K iterations on clips of resolution \(33 \times 384 \times 576\) (frames \(\times\) height \(\times\) width), requiring approximately three days. In the second stage, we train for an additional 20K iterations on longer clips of resolution \(81 \times 384 \times 576\), which takes another three days.

\mypara{Optimization} We use the AdamW optimizer with a learning rate of \(2\times10^{-5}\), weight decay of \(1\times10^{-4}\), and \(\epsilon=10^{-8}\). The learning rate follows a constant schedule with linear warmup over the first 100 steps.

\section{Benchmark Design Details}\label{appendix:benchmark}
In this section, we will introduce the benchmark design details. Table.~\ref{tab:benchmark_stats} showcases the statistic of tasks within the benchmark.
\begin{table}[!htbp]
\centering
\resizebox{.8\linewidth}{!}{
\begin{tabular}{lccc}
\toprule
\textbf{Object Type} & \textbf{Manipulation Type} & \textbf{\# Samples} \\
\midrule
\multirow{3}{*}{Ego Vehicle} 
 & Direction Change & 10 \\
 & Lane Change & 15 \\
 & Speed Change & 14 \\
\midrule
\multirow{5}{*}{Other Vehicles} 
 & Direction Change & 8 \\
 & Lane Change & 13 \\
 & Speed Change & 15 \\
 & Insertion & 14 \\
 & Removal & 20 \\
\bottomrule
\end{tabular}}
\caption{Statistics of manipulation categories in \textbf{\ourbenchmark}. 
The benchmark with in total 109 samples covers two object types (Ego vehicle and Other vehicles) and five manipulation types, 
providing diverse scene editing scenarios for controllable driving generation evaluation.}
\label{tab:benchmark_stats}
\end{table}

\subsection{Evaluation metrics formulation}\label{appendix:metrcs}

In this section, we formally describe the three task-specific controllability metrics used in \textbf{\ourbenchmark}: Vehicle Identity Matching Score (VIMS), Background Alignment Score (BAS), and Operation Success Rate (OSR).

\subsubsection{Vehicle Identity Matching Score (VIMS)}\label{appendix:vims}

VIMS measures how well the generated video preserves the identity of each vehicle instance along its trajectory, by aligning frames in 3D space and computing CLIP similarity~\cite{radford2021learning} on instance-masked regions.

\mypara{3D Frame Correspondence} For a given vehicle instance $j$, the modified trajectory provides a sequence of
vehicle-to-world transforms
\[
T^{\mathrm{mod}}_{t,j} \in SE(3), \quad t = 1,\dots,T,
\]
and ego (camera) poses
\[
E^{\mathrm{mod}}_{t} \in SE(3).
\]
We first express the vehicle pose in the ego coordinate system by composing
the transforms:
\begin{equation}
P^{\mathrm{mod}}_{t,j}
=
\bigl(E^{\mathrm{mod}}_{t}\bigr)^{-1} T^{\mathrm{mod}}_{t,j}
=
\begin{bmatrix}
R^{\mathrm{mod}}_{t,j} & p^{\mathrm{mod}}_{t,j} \\
\mathbf{0}^\top & 1
\end{bmatrix},
\end{equation}
where $R^{\mathrm{mod}}_{t,j} \in SO(3)$ and $p^{\mathrm{mod}}_{t,j} \in \mathbb{R}^3$
denote, respectively, the rotation and position of vehicle $j$ in ego
coordinates at time $t$.

Similarly, from the original trajectory we obtain vehicle-to-world transforms
$T^{\mathrm{gt}}_{k,j} \in SE(3)$ and ego poses $E^{\mathrm{gt}}_{k} \in SE(3)$
for $k = 1,\dots,K$, and we convert them to ego coordinates as
\begin{equation}
P^{\mathrm{gt}}_{k,j}
=
\bigl(E^{\mathrm{gt}}_{k}\bigr)^{-1} T^{\mathrm{gt}}_{k,j}
=
\begin{bmatrix}
R^{\mathrm{gt}}_{k,j} & p^{\mathrm{gt}}_{k,j} \\
\mathbf{0}^\top & 1
\end{bmatrix},
\end{equation}
with $R^{\mathrm{gt}}_{k,j} \in SO(3)$ and $p^{\mathrm{gt}}_{k,j} \in \mathbb{R}^3$.

Given these ego-frame poses, we define the distance between the modified pose
at time $t$ and the ground-truth pose at time $k$ as
\begin{equation}
D_{t,k,j}
=
\left\|
p^{\mathrm{mod}}_{t,j} - p^{\mathrm{gt}}_{k,j}
\right\|_2
+
0.1\,
\left\|
R^{\mathrm{mod}}_{t,j} - R^{\mathrm{gt}}_{k,j}
\right\|_F.
\end{equation}

To avoid spurious matches due to heavy occlusions, we restrict the matching to
frames in which vehicle $j$ is visible. We mark vehicle $j$ as occluded in
frame $k$ if more than $80\%$ of its projected 2D bounding box is covered by
other vehicles; otherwise it is treated as non-occluded. Let
$\mathcal{K}_{t,j} \subseteq \{1,\dots,K\}$ denote the set of ground-truth
frames in which vehicle $j$ is non-occluded. The best-matching ground-truth
frame for $(t,j)$ is then
\begin{equation}
k^\ast(t,j)
=
\arg\min_{k \in \mathcal{K}_{t,j}} D_{t,k,j}.
\label{eq:vims_frame_match}
\end{equation}

\mypara{Instance-Masked CLIP Similarity}
Let $I^{\mathrm{gen}}_t$ and $I^{\mathrm{gt}}_{k^\ast(t,j)}$ denote the generated and ground-truth frames, respectively.  
For instance $j$ at time $t$, we use its binary masks $M^{\mathrm{gen}}_{t,j}$ and $M^{\mathrm{gt}}_{k^\ast(t,j),j}$ to extract instance-only regions:
\begin{equation}
  \tilde I^{\mathrm{gen}}_{t,j} = I^{\mathrm{gen}}_t \odot M^{\mathrm{gen}}_{t,j},   
\end{equation}
\begin{equation}
\tilde I^{\mathrm{gt}}_{t,j} = I^{\mathrm{gt}}_{k^\ast(t,j)} \odot M^{\mathrm{gt}}_{k^\ast(t,j),j},
\end{equation}
where $\odot$ denotes element-wise masking. 

We then compute CLIP image embeddings $f(\cdot)$ and cosine similarity:
\begin{equation}
s_{t,j}
=
\frac{
\left\langle f\bigl(\tilde I^{\mathrm{gen}}_{t,j}\bigr),\,
        f\bigl(\tilde I^{\mathrm{gt}}_{t,j}\bigr)
\right\rangle
}{
\left\| f\bigl(\tilde I^{\mathrm{gen}}_{t,j}\bigr) \right\|_2
\left\| f\bigl(\tilde I^{\mathrm{gt}}_{t,j}\bigr) \right\|_2
}.
\end{equation}
For a single scene, we aggregate over all valid $(t,j)$ pairs:
\begin{equation}
\mathrm{VIMS}_{\text{scene}}
=
\frac{1}{N_{\text{inst}}}
\sum_{j}
\frac{1}{T_j}
\sum_{t \in \mathcal{T}_j} s_{t,j},
\label{eq:vims_scene}
\end{equation}
where $\mathcal{T}_j$ is the set of frames where instance $j$ is visible and $N_{\text{inst}}$ is the number of evaluated instances.  
In the benchmark, we report the average across scenes:
\begin{equation}
\mathrm{VIMS}
=
\frac{1}{N_{\text{scene}}}
\sum_{n=1}^{N_{\text{scene}}}
\mathrm{VIMS}_{\text{scene},n}.
\end{equation}

\subsubsection{Background Alignment Score (BAS)}\label{appendix:bas}

BAS evaluates how well the static background is preserved after editing, by explicitly masking out all dynamic instances using masks and computing CLIP similarity~\cite{radford2021learning} on the remaining background regions.

\mypara{Ego-Pose-Based Frame Matching}
For background evaluation, we match frames based on ego poses.  
Let
\[
E^{\mathrm{mod}}_t,\, E^{\mathrm{gt}}_k \in SE(3)
\]
be the ego poses at modified frame $t$ and ground-truth frame $k$, respectively.  
We define a combined distance
\begin{equation}
D(t,k)
=
\left\| x^{\mathrm{mod}}_t - x^{\mathrm{gt}}_k \right\|_2
+
0.1\,\left\| R^{\mathrm{mod}}_t - R^{\mathrm{gt}}_k \right\|_F,
\end{equation}
where $x$ is the translation component and $R$ is the $3\times3$ rotation matrix extracted from $E$, and $\|\cdot\|_F$ denotes the Frobenius norm.  
The best-matching ground-truth frame for background at time $t$ is
\begin{equation}
k^\ast(t)
=
\arg\min_{k} D(t,k).
\end{equation}

\mypara{Background Masking}
Let $M^{\mathrm{mod}}_t$ and $M^{\mathrm{gt}}_{k^\ast(t)}$ be the binary foreground masks for the generated and ground-truth videos, respectively.  
We invert them to obtain background masks:
\begin{equation}
BM^{\mathrm{mod}}_t = 1 - M^{\mathrm{mod}}_t,
\qquad
BM^{\mathrm{gt}}_{k^\ast(t)} = 1 - M^{\mathrm{gt}}_{k^\ast(t)}.
\end{equation}
We then extract background-only regions:
\begin{align}
\tilde I^{\mathrm{gen,bg}}_t &= I^{\mathrm{gen}}_t \odot BM^{\mathrm{mod}}_t, \\
\tilde I^{\mathrm{gt,bg}}_t  &= I^{\mathrm{gt}}_{k^\ast(t)} \odot BM^{\mathrm{gt}}_{k^\ast(t)}.
\end{align}

\mypara{CLIP Similarity and Aggregation}
We compute CLIP embeddings and cosine similarity:
\begin{equation}
s^{\mathrm{bg}}_t
=
\frac{
\left\langle
f\bigl(\tilde I^{\mathrm{gen,bg}}_t\bigr),\,
f\bigl(\tilde I^{\mathrm{gt,bg}}_t\bigr)
\right\rangle
}{
\left\| f\bigl(\tilde I^{\mathrm{gen,bg}}_t\bigr) \right\|_2
\left\| f\bigl(\tilde I^{\mathrm{gt,bg}}_t\bigr) \right\|_2
}.
\end{equation}
For a single scene:
\begin{equation}
\mathrm{BAS}_{\text{scene}}
=
\frac{1}{T}
\sum_{t=1}^{T} s^{\mathrm{bg}}_t,
\end{equation}
and the benchmark reports the average across all scenes.

\subsubsection{Operation Success Rate (OSR)}\label{appendix:osr}

OSR is used for \emph{insertion} and \emph{removal} tasks, and measures whether the intended operation has been successfully carried out, using GPT-4o~\cite{achiam2023gpt} as a learned judge.  
Our evaluation code follows a two-step process: (1) preparing a small set of annotated frames, and (2) querying GPT-4o for a scalar score in $[1,10]$.

\mypara{Frame Sampling and Bounding-Box Annotation}
For a given insertion or removal scene, we uniformly sample a small set of frame indices
\begin{equation}
\mathcal{S} = \{t_1,\dots,t_{K}\},
\end{equation}
typically $K=5$.  

\begin{itemize}
    \item \textbf{Insertion:} We locate the inserted object’s instance ID and load its binary mask $M^{\mathrm{ins}}_{t_k}$ on each sampled frame. We then draw a \emph{green} bounding box around the support of this mask on $I^{\mathrm{gen}}_{t_k}$.
    \item \textbf{Removal:} We locate the removed agent ID and corresponding original instance, load the mask $M^{\mathrm{rem}}_{t_k}$ for the original video, and draw a \emph{red} bounding box on the edited frame indicating where the object used to be.
\end{itemize}
The resulting annotated frames $\{\hat I_{t_k}\}_{k=1}^{K}$ are sent to GPT-4o together with a task-specific textual instruction.

\mypara{GPT-4o Prompting}
We use a concise, rating-oriented prompt.  

\begin{tcolorbox}[colback=gray!5,colframe=black,left=1mm,right=1mm,top=1mm,bottom=1mm]
\small
\textbf{Insertion prompt:}\\[2pt]
You are evaluating a generated driving video for object insertion quality.  
The goal was to insert: \textless \textit{vehicle description from metadata} \textgreater.  

The inserted object is marked with a \textbf{green} bounding box in the frames.  

Please rate the generated video on a scale of 1--10 based on:  \\
1. Realism: Does the inserted object (in green box) look realistic and well-integrated? \\
2. Consistency: Is the object consistent across frames? \\
3. Alignment: Does it match the text description? \\
4. Physical plausibility: Does it follow realistic motion and positioning? \\
5. Visual quality: Are there any artifacts or inconsistencies in the boxed region? 

Provide ONLY a single number from 1 to 10 as your response.
\end{tcolorbox}

\begin{tcolorbox}[colback=gray!5,colframe=black,left=1mm,right=1mm,top=1mm,bottom=1mm]
\small
\textbf{Removal prompt:}\\[2pt]
You are evaluating a generated driving video for object removal quality.  
The goal was to remove: \textless \textit{vehicle description from metadata} \textgreater.  

The inserted object is marked with a \textbf{red} bounding box in the frames.  

Please rate the generated video on a scale of 1--10 based on:  \\
1. Completeness: Is the object fully removed from the red box area? \\
2. Naturalness: Does the background look natural where the object was (in red box)? \\
3. Consistency: Is the removal consistent across frames? \\
4. Artifacts: Are there visible artifacts or inconsistencies in the boxed region? \\
5. Inpainting quality: How well is the removed region filled in? 

Provide ONLY a single number from 1 to 10 as your response.
\end{tcolorbox}

GPT-4o returns a string which is parsed as a scalar score
\begin{equation}
g_i \in [1,10]
\end{equation}
for scene $i$.  

\mypara{Aggregation}
We define OSR as the average GPT score over all scenes:
\begin{equation}
\mathrm{OSR}
=
\frac{1}{N_{\text{ops}}}
\sum_{i=1}^{N_{\text{ops}}} g_i,
\end{equation}
where $N_{\text{ops}}$ is the number of evaluated insertion or removal scenes.  
This gives a human-aligned, continuous measure of operation success on a 1--10 scale.

\section{Downstream Tasks}\label{appendix:down}

To evaluate the utility of our generated data for downstream perception, we conduct a 3D object detection study using BEVFormer~\cite{li2024bevformer} on the Waymo Open Dataset~\cite{sun2020waymo}. Following the protocol of ChatSim~\cite{wei2024editable}, we mix the real training set with simulated data and compare performance before and after mixing. We start from $8000$ real training frames and generate $8000$ corresponding simulated versions. For each sequence, we produced two edited variants with our framework:  
(i) \textbf{Insertion} (\textbf{Ins}): randomly inserting four novel vehicles as replacements for existing ones;  
(ii) \textbf{Removal} (\textbf{Rem}): randomly removing vehicles from the scene.  
Each operation contributes an additional $8000$ edited frames.

We evaluate BEVFormer on the full Waymo validation set using mean Average Precision (AP) at IoU thresholds of 0.3, 0.5, and 0.7.

\begin{table}[!htbp]
\centering
\resizebox{\linewidth}{!}{
\begin{tabular}{lccc}
\toprule
\textbf{Training Data} & \textbf{AP@0.3} & \textbf{AP@0.5} & \textbf{AP@0.7} \\
\midrule
Org & 0.1211 & 0.0533 & 0.0103 \\
Org + Ins & 0.1343 & \cellcolor{green!15}\textbf{0.0635} & 0.0125 \\
Org + Rem & \cellcolor{green!15}\textbf{0.1359} & 0.0613 & \cellcolor{green!15}\textbf{0.0134} \\
\bottomrule
\end{tabular}}
\caption{Downstream 3D detection results of BEVFormer~\cite{li2024bevformer} on the Waymo validation set under a front-camera-only configuration. \textbf{Org} denotes the original $8000$ real frames. \textbf{Ins} and \textbf{Rem} denote additional frames generated using our method with random vehicle insertion and removal, respectively.}
\label{tab:supp_downstream}
\end{table}

As shown in Table~\ref{tab:supp_downstream}, augmenting the training set with our edited scenes leads to \emph{substantial} relative gains across all IoU thresholds. 
Adding insertion-based edits (\textbf{Org + Ins}) improves AP from $0.1211$ to $0.1343$ at IoU~0.3 (\textbf{+10.9\%} relative), from $0.0533$ to $0.0635$ at IoU~0.5 (\textbf{+19.1\%}), and from $0.0103$ to $0.0125$ at IoU~0.7 (\textbf{+21.4\%}). 
Removal-based edits (\textbf{Org + Rem}) provide similarly strong or even larger gains, boosting AP to $0.1359$, $0.0613$, and $0.0134$ at IoU~0.3/0.5/0.7, corresponding to \textbf{+12.2\%}, \textbf{+15.0\%}, and \textbf{+30.1\%} relative improvements over the \textbf{Org} baseline. 
These large relative gains demonstrate that our generated data is not only visually realistic but also highly informative, and that controllable scene editing with our framework can effectively enrich training distributions for autonomous driving perception.

\section{User Study Design}\label{appendix:user}
In this section we will introduce the design of our user study.
For every sample, participants were provided with: (1) the original video and trajectory visualization before editing, (2) the target trajectory visualization specifying the desired vehicle behavior and (3) four edited results from different methods, denoted as methods A, B, C, and D. The participant’s task was to select the one video that best aligns with the target trajectory and maintains the highest visual realism. To ensure fairness and eliminate potential bias or cherry-picking, the sampling and ordering were fully randomized: each participant received a unique questionnaire where both the 20 selected samples and the order of methods (A–D) were shuffled independently. This guarantees that no two participants saw identical question sets, and that every video sample was equally likely to appear across users.

\section{Fixed-Offset Shifting Performance}
We further evaluate fixed-offset shifting, as shown in Table~\ref{tab:lane_shift}, 
where our method notably outperforms ReconDreamer~\cite{ni2025recondreamer} and DriveDreamer4D~\cite{zhao2024drivedreamer4d} in NTA-IoU, NTL-IoU and FID, proving the superior positional control accuracy and robustness of the proposed method.
\begin{table}[htbp] %
    \centering
    \resizebox{1.0\linewidth}{!}{
        \begin{tabular}{lcccccc}
        \toprule
        \multirow{2}{*}{\textbf{Method}} & \multicolumn{2}{c}{\textbf{NTA-IoU} ($\uparrow$)} & \multicolumn{2}{c}{\textbf{NTL-IoU} ($\uparrow$)}& \multicolumn{2}{c}{\textbf{FID} ($\downarrow$)}\\ \cline{2-3}  \cline{4-5} \cline{6-7} 
         & \textbf{Lane Shift @ 3m} & \textbf{Lane Shift @ 6m} & \textbf{Lane Shift @ 3m} & \textbf{Lane Shift @ 6m} & \textbf{Lane Shift @ 3m} & \textbf{Lane Shift @ 6m}\\ \midrule
         \textbf{DriveDreamer4D} & 0.340 & 0.121 & 51.32 & 49.28 & 129.05 & 210.37\\
        \textbf{ReconDreamer} & 0.539 & 0.467 & 54.58 & 52.58& 93.56 & 149.19\\
        \textbf{Ours} & \textbf{0.552} & \textbf{0.523} &\textbf{56.73}& \textbf{54.01}& \textbf{49.47} & \textbf{53.61}\\ \bottomrule
        \end{tabular}
    }
    \caption{Comparison on large-offset manipulation.}
    \label{tab:lane_shift} 
\end{table}

\section{Challenging Inputs Performance}
\begin{figure}[htbp]
    \centering
    \includegraphics[width=1.0\linewidth]{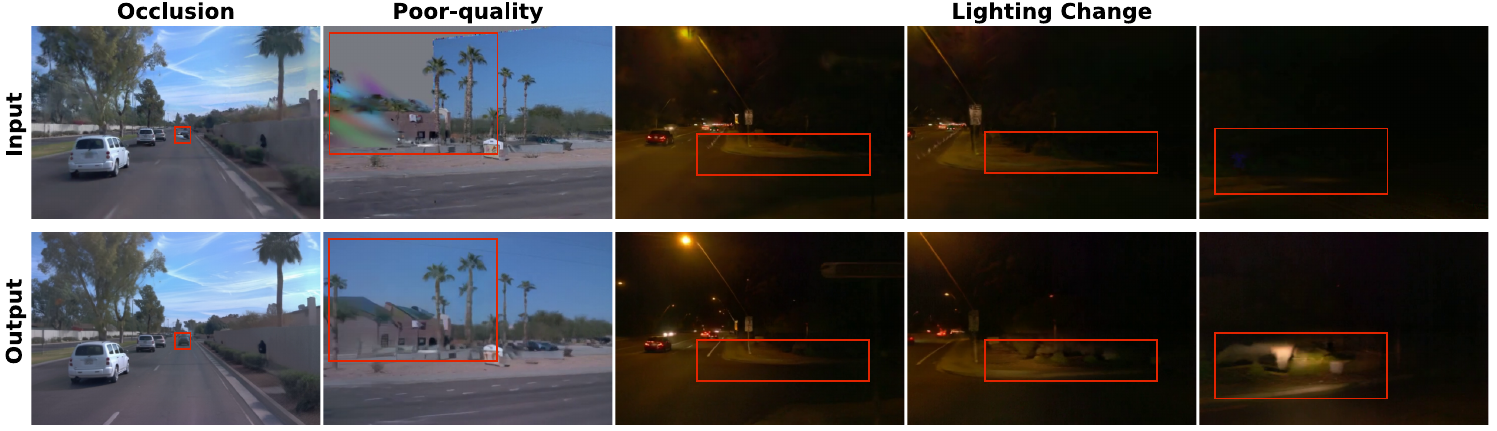} 
    \caption{Visualization under challenging conditions. 
    }
    \label{fig:challenging_cases}
\end{figure}

As shown in Figure~\ref{fig:challenging_cases}, our method demonstrates robust performance under challenging inputs: 
correcting vehicle artifacts caused by partial occlusion in original views, addressing severe sky artifacts, and naturally rendering dynamic headlight illumination during nighttime turns—proving our capability to handle poor-quality inputs, heavy occlusion, or dynamic background.

\section{OSR Reliability}
We validated OSR reliability through 
a human evaluation using the identical 1-10 rubric.
As shown in Table~\ref{tab:osr_human}, this alignment confirms OSR serves as a reliable metric. 

\begin{table}[htbp]
    \centering
    \resizebox{\linewidth}{!}{
        \begin{tabular}{l|cc|cc|cc|cc}
        \toprule
        \multirow{2}{*}{\textbf{Metric}} & \multicolumn{2}{c|}{\textbf{Ours}} & \multicolumn{2}{c|}{\textbf{Difix3D}} & \multicolumn{2}{c|}{\textbf{OmniRe}} & \multicolumn{2}{c}{\textbf{StreetCrafter}} \\ \cline{2-9} 
         & \textbf{Ins.} & \textbf{Rem.} & \textbf{Ins.} & \textbf{Rem.} & \textbf{Ins.} & \textbf{Rem.} & \textbf{Ins.} & \textbf{Rem.} \\ \midrule
        \textbf{OSR} & \textbf{5.86} & \textbf{7.00} & 4.71 & 6.89 & 4.23 & 6.42 & 4.62 & 6.51 \\
        \textbf{Human} & \textbf{7.50} & \textbf{7.25} & 3.79 & 4.40 & 3.58 & 4.75 & 3.12 & 3.25 \\ \bottomrule
        \end{tabular}
    }
    \caption{Comparison of OSR and Human Scores.}
    \label{tab:osr_human}
\end{table}

\section{More visualization results}

For more visualization results, please check the webpage.

\end{document}